\documentclass[letterpaper]{article} 
\usepackage{aaai2026}  
\usepackage{times}  
\usepackage{helvet}  
\usepackage{courier}  
\usepackage{amsmath}  
\usepackage{amssymb}  
\usepackage{mathtools}  
\usepackage{booktabs}  
\usepackage{multirow}  
\usepackage{subcaption}  
\usepackage{tcolorbox}  
\usepackage{enumitem}  
\usepackage[hyphens]{url}  
\usepackage{graphicx} 
\usepackage{natbib}  
\usepackage{caption} 
\usepackage{algorithm}
\usepackage{algorithmic}

\usepackage{newfloat}
\usepackage{listings}
\DeclareCaptionStyle{ruled}{labelfont=normalfont,labelsep=colon,strut=off} 
\floatstyle{ruled}
\newfloat{listing}{tb}{lst}{}
\floatname{listing}{Listing}
\title{Bridging Scale Discrepancies in Robotic Control via Language-Based Action Representations}
\author{
    Yuchi Zhang\textsuperscript{\rm 1},
    Churui Sun\textsuperscript{\rm 1},
    Shiqi Liang\textsuperscript{\rm 1},
    Diyuan Liu\textsuperscript{\rm 2},
    Chao Ji\textsuperscript{\rm 2},
    Wei-Nan Zhang\textsuperscript{\rm 1,3}\thanks{Corresponding author.},
    Ting Liu\textsuperscript{\rm 1}
}
\affiliations{
    \textsuperscript{\rm 1}Research Center for Social Computing and Interactive Robotics, Harbin Institute of Technology, Harbin, China\\
    \textsuperscript{\rm 2}State Key Laboratory of Cognitive Intelligence, iFLYTEK Research, China\\
    \textsuperscript{\rm 3}Suzhou Research Institute, Harbin Institute of Technology, Suzhou, China\\
    \{yczhang, crsun, sqliang, wnzhang, tliu\}@ir.hit.edu.cn\\
    \{dyliu2, chaoji\}@iflytek.com
}

\usepackage{bibentry}

\begin{document}

\maketitle

\begin{abstract}
Recent end-to-end robotic manipulation research increasingly adopts architectures inspired by large language models to enable robust manipulation. However, a critical challenge arises from severe distribution shifts between robotic action data, primarily due to substantial numerical variations in action commands across diverse robotic platforms and tasks, hindering the effective transfer of pretrained knowledge. To address this limitation, we propose a semantically grounded linguistic representation to normalize actions for efficient pretraining. Unlike conventional discretized action representations that are sensitive to numerical scales, the motion representation specifically disregards numeric scale effects, emphasizing directionality instead. This abstraction mitigates distribution shifts, yielding a more generalizable pretraining representation. Moreover, using the motion representation narrows the feature distance between action tokens and standard vocabulary tokens, mitigating modality gaps. Multi-task experiments on two benchmarks demonstrate that the proposed method significantly improves generalization performance and transferability in robotic manipulation tasks.
\end{abstract}

\section{Introduction}

Recent advances in artificial intelligence have enabled models to acquire extensive knowledge from large-scale data, demonstrating robust generalization across tasks with minimal fine-tuning \citep{radford2019language,brown2020languagemodelsfewshotlearners}. Building on foundational robotics research that integrated vision and control\citep{chaumette2006visual,saxena2008robotic}, modern approaches increasingly adapt these large-scale AI methods to robotics \citep{Surveyyifanchen} . In particular, language-conditioned action learning leverages both visual and linguistic inputs to guide robot actions, enabling more flexible and versatile manipulation capabilities.
Large-scale datasets like Open X-Embodiment (OXE) \citep{10611477}aggregate multimodal robotic demonstrations across 22 robot embodiments and over 1 million tasks. By unifying visual (RGB/depth), proprioceptive, and language inputs with action trajectories in a standardized format, OXE facilitates cross-robot policy learning. Advanced works like OpenVLA, Octo, $Pi_0$, and RDT \citep{kim2024openvla,team2024octo, black2410pi0,liu2024rdt} leverage these datasets to explore model architectures and improving robotic manipulation.

\begin{figure*}[t] 
\begin{center}
\includegraphics[width=0.9\linewidth]{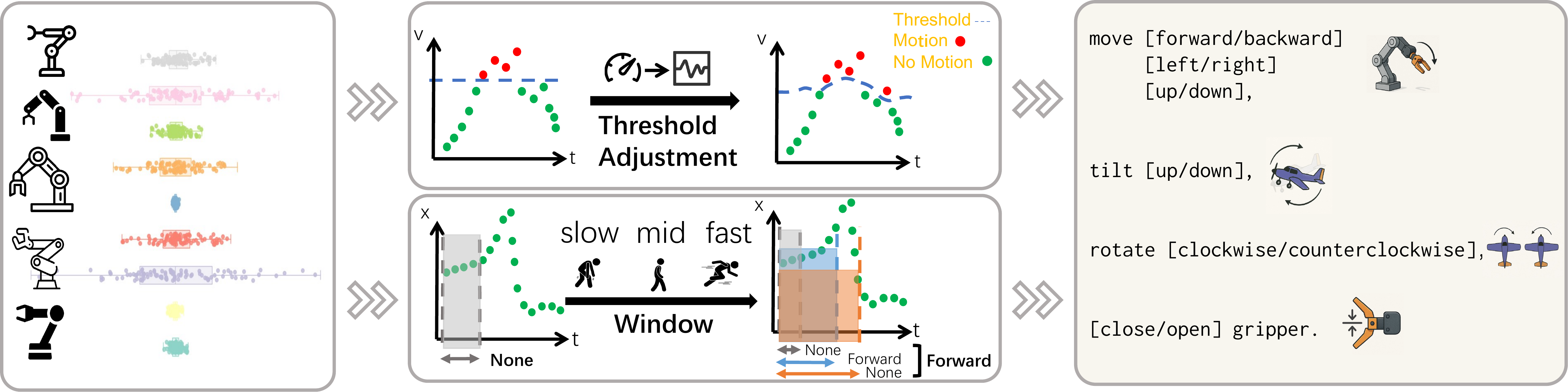} 
\end{center}
\caption{The proposed motion data generation pipeline. The left part illustrates the distributions of specific execution actions across different types of datasets; The middle part presents our threshold- and window-based detection framework along with its proposed improvements; The right part depicts the structure and representation of the generated motion outputs.}
\label{fig:main_figure} 
\end{figure*}

Despite the successes in NLP through autoregressive pretraining, robotics models still face significant challenges in achieving similar transferable generalization. One of the main obstacles is distribution shifts across datasets caused by variations in collection environments, visual conditions, robot hardware, and data collection protocols. Consequently, current models frequently require extensive fine-tuning to perform satisfactorily in new domains.
Additionally, existing language-conditioned imitation learning approaches often provide dynamic visual inputs at each timestep but maintain static language instructions. This imbalance limits the influence of language modality in guiding action generation, failing to fully utilize language's potential.

To address the above limitations, we introduce a language-based intermediate representation to guide robot actions before execution, achieved through a rule-based mapping that transforms end-effector actions into coarse-grained language descriptions as alignment targets. To handle distribution shifts and ensure adaptability across diverse datasets, the proposed method incorporates a generalized motion generation method that applies spatial normalization and dynamic threshold adjustments. Leveraging the semantic richness and robust generalization of natural language, our unified pretraining strategy autonomously generates accurate, cost-efficient language alignment targets from diverse datasets without relying on external modules or manual intervention, thereby enhancing generalization and adaptability. We first train on a subset of the OXE dataset to capture execution patterns across varied environments, then fine-tune the model on manipulation benchmark datasets under language-alignment constraints to ensure semantic correspondence between actions and language. Experiments demonstrate that this approach enhances transferability, execution accuracy, and stability, with language alignment further bolstering robustness under diverse task conditions.

In summary, our contributions include:
\begin{itemize}[leftmargin=*]
    \item We propose a novel pretraining strategy leveraging rule-based linguistic representations to align action-language distributions across datasets, inherently capturing generalized motion-language relationships without manual annotations or external corrections, thus enhancing model generalization and transferability.
    \item We propose an adaptive multi-scale motion detection method that dynamically adjusts thresholds and employs hierarchical windows, effectively suppressing motion jitter and false segmentation across datasets, significantly improving the accuracy of complex action recognition.
    \item Extensive evaluations on LIBERO \citep{liu2023liberobenchmarkingknowledgetransfer} and Simpler Env \citep{li24simpler} benchmarks validate our method's superior accuracy, stability, and robustness compared to existing baselines.
\end{itemize}

\section{Related Work}

\subsection{End-to-End Action Generation }
In the manipulation field, there are many attempts to train models on a large scale end-to-end. Typical examples include RT1\citep{brohan2022rt} and RT2\citep{brohan2023rt}, which use FiLM\citep{perez2018film} and CLIP\citep{radford2021learning} for image encoding, transformer as the backbone, and discrete action space instead of act token for action decoding. RT2\citep{brohan2023rt} also uses large-scale mixed data to allow the model to perform actions while retaining some multimodal QA knowledge. Other work from the same period includes Octo\citep{team2024octo}, which also uses Transformer as the backbone and is pretrained on the largest robot manipulation dataset Open X-Embodiment. Moreover, there are many similar works, including Openvla\citep{kim2024openvla} and $Pi_0$\citep{black2410pi0}. Some work has noticed that when using multiple robot arm data for pretraining, there will be issues of embodiment inconsistency. To solve this problem, RDT\citep{liu2024rdt} introduces a Physically Interpretable Unified Action Space to unify data from different sources, while HPT\citep{wang2024scaling} utilizes embodiment-specific tokenizers (``stems"), mapping the proprioception and visual sensing information of different robotic arms into a shared latent space.

\subsection{Action Generation Assisted by Textual Guidance}
Some researchers believe that in manipulation tasks, semantic expressions are becoming more diverse, making the mapping from high-level tasks to specific operations more difficult. To address this issue, some research proposes letting the model first learn the mapping from tasks to general language descriptions and then further learn specific operational actions. However, the model may have biases when generating actions based on language descriptions. Therefore, RT-H\citep{belkhale2024rt} introduced a manual intervention mechanism to correct errors in language descriptions, while ECoT\citep{zawalski2024robotic} extended the language reasoning chain to guide correct action descriptions, exploring the effectiveness of ChatGPT in correcting actions. Additionally, Emma\citep{sun2024emma} further improved chain-of-thought generation and introduced explicit state information from trajectories as input to enhance the model's task understanding and execution capabilities. Similarly, CoA\citep{li2024improving} proposed Chain-of-Affordance, using the location of affordances in images as a chain of thought to guide the model in generating more robust actions. Meanwhile, some work\citep{qi2025sofar} considers object orientations to be a key requirement for fine-grained manipulations tasks; They constructed a dataset of object-text-orientation pairs, excelling in many embodied tasks. In contrast to the above methods that rely on explicit guidance, we propose enhancing the model's action generation capability through multi-dataset pretraining. This approach enables the model to produce more robust motion language descriptions, thereby significantly improving its generalization performance across tasks and datasets—without requiring additional reasoning chains or manual correction.  \citep{fu2024msi} .

\section{Methods}

This section presents our method in the order of action tokenization, motion generation, and model training. First, \textit{Action Tokenizer} discretizes continuous action signals into token sequences to establish a learnable output space. Second, \textit{Motion Generation} generates robust natural-language motion signals using adaptive thresholds and hierarchical temporal windows, serving as high-level semantic guidance. Finally, the \textit{Two-Stage Training} stage employs a two-stage conditional generation strategy to progressively predict concrete actions from observations and instructions.

\begin{figure*}[!t] 
\begin{center}
\includegraphics[width=0.75\linewidth]{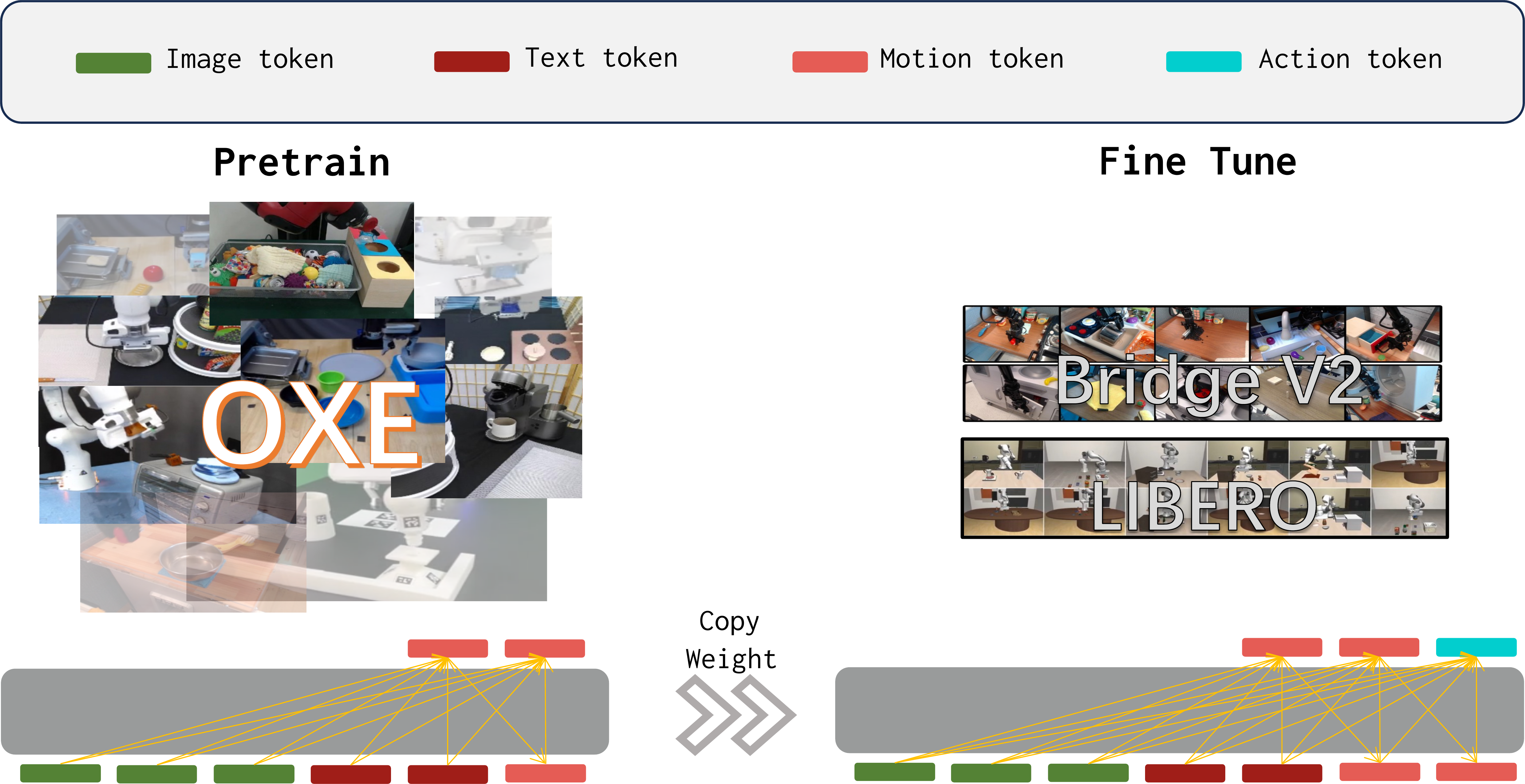} 
\end{center}
\caption{Two-stage training on Qwen2.5 (0.5B, 1.5B, 3B): pretraining predicts motion tokens; fine-tuning predicts motion then action tokens. Image tokens denote the observed visual input; text tokens denote the task instruction; motion tokens denote our proposed motion language; and action tokens denote the discrete action representation. }
\label{fig:train} 
\end{figure*}

\subsection{Action Tokenizer}
Our action decoding method follows the approach of RT-2 and OpenVLA. Based on the task instruction and current observation input, the VLA should predict 7 action tokens consecutively, representing a 7-dimensional action the robot should execute, with each dimension corresponding to $(\Delta X, \Delta Y, \Delta Z, \Delta \text{roll}, \Delta \text{pitch}, \Delta \text{yaw}, \text{GripperState})$.
The variables are normalized during training, and the output is denormalized during inference. Each normalized variable is discretized into 256 bins, where each bin is represented as a unique token. This transforms the action prediction task into a token-based sequence prediction task. 
In our VLA design, we appended 256 additional tokens to the tail, specifically to represent the 256 action tokens, denoted as \verb|<extra_0>|--\verb|<extra_255>|.
For normalizing, we exclude the outliers in each of the seven dimensions that fall outside the 1st and 99th percentile range. If outliers are included, the normalization range expands significantly, resulting in coarse-grained predictions and larger bin sizes, which can negatively impact precision.

\subsection{Motion Generation}
Previous work typically relies on manually defined thresholds and window sizes to generate motion signals. The \textbf{threshold} distinguishes actions by treating motion magnitude above it as an active movement and below it as no action in that dimension, while the \textbf{window} defines the temporal span over which motion displacements are accumulated into a single motion token. While such methods perform reasonably well on individual datasets, their simplicity makes them poorly suited for handling complex motion patterns across multiple datasets.

To enable collaborative motion generation in a multi-dataset setting, the normalization method described in subsection \textit{Action Tokenizer} is first applied before generating motion signals. Building upon this, we account for the jittering phenomenon commonly observed in robotic arms operating in real-world environments by replacing fixed thresholds with adaptive ones. Additionally, to accommodate the diverse types of robotic arm movements, we replace the single fixed-size window with a hierarchical detection window.
\subsubsection{Motion Representation}
The ``motion" representations we generated are a fixed set of natural language descriptions structured as follows:
move [forward/backward] [left/right] [up/down],
tilt [up/down],
rotate [clockwise/counterclockwise],
[open/close] gripper.
Specifically, the keyword move describes positional displacement of the actuator along coordinate axes, while tilt and rotate denote angular rotations of the actuator, and gripper refers to its open-close actions. Each ``motion" representation is composed of a combination of these three movement types. In cases where no movement is detected across all dimensions, the motion token is labeled as ``stop".

\subsubsection{Threshold} The threshold defines the minimum motion magnitude required to consider a change as meaningful—motion below this threshold is treated as no action in the corresponding dimension.
To achieve more precise threshold determination for motion detection, we specifically take into account the jittering phenomena that may arise from faster movement speeds. We introduce a speed-based correction method, which adjusts the threshold to compensate for jitter caused by high-speed motion.

Let $T_{base}^i$ denotes the basic threshold, and $\beta$ represents the sensitivity coefficient. Additionally, $\tau$ represents the threshold adjustment window. The detecting formula is shown below, where $s$ indexes the time steps within the window:
\begin{equation}
T_i(t) = T_{base}^i + \beta\cdot\frac{1}{\tau}\sum_{t-\tau}^t|\hat{\Delta}_i(s)|
\end{equation}

\subsubsection{Window} Since motion generation often involves accumulation over multiple frames, the window determines the temporal span over which this accumulation occurs. Oriented to fixed window for motion detecting, incorporating the approach of fast-varying subsystem and slow-varying subsystem in singular systems, we designed the layered detecting window for motion categories comprising three temporal resolutions: fast, mid, and slow. 
Let \( p(t) \in \mathbb{R}^3 \) denote the 3D gripper position at time \( t \), and let \( T \in \mathbb{R}^+ \) be a predefined movement threshold.
To simplify expressions, we define the following shorthand notations:
\begin{itemize}
    \item \( \Delta_t p \coloneqq p(t) - p(t - 1) \): unit-step displacement at time \( t \)
    \item \( \Delta_{\text{t}_X} p \coloneqq p(t) - p(t - \Delta t_{\text{X}}) \): displacement over a window of size \( \Delta t_{\text{X}} \), where \( \text{X} \in \{\text{fast}, \text{mid}, \text{slow}\} \)
\end{itemize}
For brevity, we use subscripts f, m, s to denote fast, mid, slow respectively in the following definitions. The motion detectors for each temporal level are then defined as follows:
\begin{equation}
    \text{M}_{\text{f}}\coloneqq \|\Delta_{\text{t}_{f}} p\| > 2T
\end{equation} 
Focusing on regular movements, we modified the motion judging logic based on ECOT\citep{zawalski2024robotic}, ensuring the robotic arm is always in a moving process. 


\begin{equation}
\text{M}_\text{m} \coloneqq 
  \|\Delta_{\text{t}_{m}} p\| > T
  \;\wedge\;
  \min_{\tau \in [t - \Delta t_{\text{m}},\, t]} \|\Delta_\tau p\| > 0
\end{equation}

Focusing on the slow reactions in slow-varying systems, considering that these reactions usually have long response times and slow-changing state variables, we designed a threshold detection under a large window. Meanwhile, to avoid multiple segments of motion being recognized as a whole slow reaction due to an overly large window, we stipulated that the motion must proceed steadily in the same direction.

\begin{equation}
\text{M}_\text{s}\coloneqq
  \|\Delta_{\text{t}_\text{s}} p\| > T \wedge 
  \min_{\tau \in [t{-}\Delta t_{\text{s}},\,t]} \|\Delta_\tau p\| > \frac{T}{2\Delta t_{\text{s}}}
\end{equation}


We comprehensively evaluate the actions, identifying them only after each action detection has been passed.
\begin{equation}
\text{Motion}(t) \coloneqq 
\text{M}_{\text{f}}(t) \vee \text{M}_{\text{m}}(t) \vee \text{M}_{\text{s}}(t)
\end{equation}

\subsubsection{Design Justification}
To compare the advantages and disadvantages of our proposed method with the fixed threshold-based approach used in ECoT, we manually annotated 5\% of the data (or 3\% for some larger datasets) and computed the average accuracy of action annotations. Given the distributional differences across datasets, we used separately tuned thresholds for each dataset when applying the baseline method.

The average annotation accuracy of our method reached 86.37\%, significantly outperforming the ECoT-style threshold method, which achieved only 57.62\%.

Upon further analysis of the failure cases, we observed that the threshold-based method—due to its simplicity—tends to falsely identify minor jitter during execution as multiple distinct actions. In contrast, our method effectively suppresses such false positives, leading to more stable and robust action recognition across different datasets.

\subsection{Two-Stage Training}

For each manipulation trajectory $i$, we associate a task instruction, formulated as: \texttt{"What action should the robot take to \{instruction\}?"}. A trajectory consists of a sequence of discrete actions $A_i = (a_i^0,a_i^1,\dots,a_i^T)$, where $T$ denotes the total number of steps in the trajectory. These actions align with video frames captured from a certain viewpoint, denoted $O_i = (o_i^0,o_i^1,\dots,o_i^T)$. Here, we employ third-person video frames as the image input and fix the temporal window to 1, leading each trajectory to produce one data instance per step.
To enable an end-to-end cross-modal mapping, we introduce a language-based “motion” modality $M_i = (m_i^0,m_i^1,\dots,m_i^T)$ in the output to describe the actions. These coarse-grained labels serve as an intermediate representation, ultimately forming tuples of the form $(O_i^j , p_i , M_i^j, A_i^j)$, where $i$ indexes the trajectory, $j$ indexes the step within trajectory $i$, and $p_i$ denotes the task instruction associated with the $i$-th trajectory.
The strategy we aim to learn consists of two stages: firstly, $\phi_h(m|o, p)$ indicates that, conditioned on the current observation and instruction, the model generates motion tokens describing the upcoming actions in an autoregressive, next-token-prediction manner. Subsequently, $\phi_l(a|o, p, m)$ leverages these predicted motion tokens as contextual information to infer specific action tokens.
\begin{equation}
\phi(a, m|o, p) = \phi_h(m|o, p)\phi_l(a|o, p, m)
\end{equation}

\noindent\textbf{Motion-Only Pretraining}
\begin{table*}[t]
\begin{tcolorbox}[colback=gray!10!white, colframe=black!50!white, boxrule=0.8pt, arc=5pt]
\begin{tabbing}
\hspace{1.5em}\= \kill 

{\textless START\textgreater}system\textbackslash n   $\mathbf{X}_{\text{system}}$  \textcolor{green!50!black}{\textless STOP\textgreater} \textbackslash n \\
{\textless START\textgreater}user\textbackslash n   What action should the robot take to $\mathbf{X}_{\text{instruct}}$ ? \textcolor{green!50!black}{\textless STOP\textgreater} \textbackslash n \\
{\textless START\textgreater}motion\textbackslash n \textcolor{green!50!black}{$\mathbf{X}_{motion}$} \textcolor{green!50!black}{\textless STOP\textgreater} \textbackslash n

\end{tabbing}
\end{tcolorbox}

\caption{The input sequence used to pretrain the model. In practice. In our current implementation, $\mathbf{X}_{\text{system}} = $ \texttt{You are Qwen, created by Alibaba Cloud. You are a helpful assistant.} and \textless START\textgreater = \texttt{\textless|im\_start|\textgreater},\textcolor{green!50!black}{\textless STOP\textgreater} = \texttt{\textless|im\_end|\textgreater}. The model is trained to predict the robot motion and where to stop, and thus only \textcolor{green!50!black}{green sequence/tokens} are used to compute the loss in the auto-regressive model.}
\end{table*}
We believe that the VLA model during pretraining encounters multi-source data with significantly different distributions, where action offsets are particularly severe, caused by the sampling frequencies and entity machine differences at the time of dataset construction. Therefore, existing pretraining often struggles to capture general information. In contrast, the motions generated on each dataset, as described in the previous subsection, are relatively more unified. Following the idea similar to curriculum learning, we start with easier tasks, so we hope that the pretraining phase can more efficiently capture general directional knowledge. Meanwhile, the difficulty of both learning and transfer will be greatly reduced, aligning with the principles of curriculum learning \citep{qi2024interactive} .

The format of the training data follows the construction method of VLM supervised fine-tuning data (llava), as presented in Table 1.

\noindent\textbf{Downstream Fine-Tuning}
\begin{table*}[t]
\begin{tcolorbox}[colback=gray!10!white, colframe=black!50!white, boxrule=0.8pt, arc=5pt]
\begin{tabbing}
\hspace{1.5em}\= \kill 

{\textless START\textgreater}system\textbackslash n   $\mathbf{X}_{\text{system}}$  \textcolor{green!50!black}{\textless STOP\textgreater} \textbackslash n \\
{\textless START\textgreater}user\textbackslash n   What action should the robot take to $\mathbf{X}_{\text{instruct}}$ ? \textcolor{green!50!black}{\textless STOP\textgreater} \textbackslash n \\
{\textless START\textgreater}motion\textbackslash n \textcolor{green!50!black}{$\mathbf{X}_{motion}$} \textcolor{green!50!black}{\textless STOP\textgreater} \textbackslash n
\\
{\textless START\textgreater}assistant\textbackslash n \textcolor{green!50!black}{$\mathbf{X}_{action}$} \textcolor{green!50!black}{\textless STOP\textgreater} \textbackslash n

\end{tabbing}
\end{tcolorbox}

\caption{Most settings are the same as in the previous subsection, but in this period, the model needs to predict both motion and robotic arm actions simultaneously.}
\end{table*}
After a broad and diverse pretraining phase, we expect the model to acquire more general motion representations. However, due to data distribution shifts, none of the existing VLA models can achieve zero-shot generalization and need further adaptation through imitation learning data from downstream scenarios. Therefore, fine-tuning is an essential process.

Meanwhile, since our pretraining objective is to learn more general representations, although this motion also represents an action, it is too coarse-grained. As pointed out by \citep{li2024towards}, coarse-grained predictions are easier to make under the same conditions but perform far worse in execution compared to fine-grained ones. Therefore, we need finer-grained action tokens to represent the actions to be executed. The training data follows the construction format of VLM supervised fine-tuning data, as presented in Table 2.



\section{Experimental Setup}

\subsection{Research Questions}
In our experimental evaluation, we aim to answer the following questions:

\begin{enumerate}[leftmargin=*,label=\textbf{(RQ\arabic*)}]
    \item What is the individual contribution of each refinement on performance?
    \item Does the model with these refinements outperform established baselines and state-of-the-art approaches?
    \item Does adding a language output objective reduce the gap between action tokens and language tokens?
\end{enumerate}

\subsection{Baseline Methods}

We compare with recent baselines including \textbf{Diffusion Policy}~\citep{chi2023diffusionpolicy}, \textbf{ScaleDP}~\citep{zhu2024scalingdiffusionpolicytransformer}, \textbf{Octo}~\citep{team2024octo}, \textbf{OpenVLA}~\citep{kim2024openvla}, \textbf{RT-1-x}~\citep{brohan2022rt}, and \textbf{ECoT}~\citep{zawalski2024robotic}, covering diffusion-based, vision-language-action, and transformer-based policies. Details are provided in Appendix A. 

\subsection{Dataset}
During pretraining, we utilized Open X-Embodiment, a large and diverse dataset containing hundreds of thousands of demonstrations. To reduce the computational cost during the pretraining phase, we selected 7 sub-datasets from it, including furniture-bench\citep{heo2023furniturebench} and jaco\citep{dass2023jacoplay}, totalling approximately 12,000 trajectories. More details are in the Appendix B. This amount of data has shown promising results in demonstrating the benefits of motion pre-training in our experiments. To evaluate the generalization ability of the pretraining method, our pretraining dataset does not include LIBERO and Bridge V2\citep{walke2024bridgedatav2datasetrobot}. Based on this, we generated motion data for pretraining using the pipeline introduced in Section Method.

For fine-tuning, we applied the same pipeline to LIBERO and Bridge V2 datasets. LIBERO features 130+ language-conditioned manipulation tasks for studying knowledge transfer in lifelong learning. Bridge V2 includes 7,200 demonstrations spanning 10 environments and 71 tasks in household scenarios.




\subsection{Evaluation}
\textbf{LIBERO} is a benchmark of 130+ language-conditioned tasks for Lifelong Decision-Making Learning (LLDM), focusing on knowledge transfer and skill personalization. We tested on four suites: Spatial, Goal, Object, and Long, following the open-source OpenVLA settings.

\noindent\textbf{SimplerEnv} offers a scalable simulation for real-world robot manipulation. For Bridge v2, SimplerEnv evaluates success rates on four tasks: Place a spoon on a towel, Place a carrot on a plate, Stack a green block on a yellow block and Place an eggplant in a yellow basket.




\subsection{Implementation Details} 
{\bf Model Architecture } 
Our architecture builds on OpenVLA, standardizing image resolution to 224 × 224px and encoding with SigLIP\citep{zhai2023sigmoidlosslanguageimage} and DINO v2\citep{oquab2024dinov2learningrobustvisual}, followed by channel-wise concatenation. The LLM backbone uses Qwen2.5\citep{qwen2025qwen25technicalreport} in three sizes: 0.5B, 1.5B, and 3B, with 256 special tokens added to the action tokenizer for 256 bins.

We replicated two-stage VLM training for Qwen2.5 using the LLaVA 1.5 data mixture in the Prismatic\citep{karamcheti2024prismaticvlmsinvestigatingdesign} framework, then trained using our two-stage method. To verify the impact of pretraining, we conducted experiments from scratch, performing direct fine-tuning.

\noindent{\bf Fine-tuning Hyperparameters} For pretraining, we used a batch size of 2048, and for fine-tuning, 512, with a learning rate of 2e-5. All experiments were conducted on A100-80G GPUs. For detailed information regarding the training duration, inference time, and computational efficiency, as well as the time spent on each experimental phase and reasoning process, please refer to Appendix C.

\section{Experimental Results and Analysis}
\subsection{How does each refinement impact performance individually? (RQ1)}

\begin{table}[t]
\centering
\scriptsize
\setlength{\tabcolsep}{1.5pt} 
\renewcommand{\arraystretch}{1.1} 
\begin{tabular}{cc|c|c|c|c|c|c}
\toprule
\multicolumn{2}{c|}{\textbf{From Scratch}} & 
\begin{tabular}[c]{@{}c@{}}\textbf{LIBERO-}\\\textbf{Spatial}\end{tabular} &
\begin{tabular}[c]{@{}c@{}}\textbf{LIBERO-}\\\textbf{Object}\end{tabular} &
\begin{tabular}[c]{@{}c@{}}\textbf{LIBERO-}\\\textbf{Goal}\end{tabular} &
\begin{tabular}[c]{@{}c@{}}\textbf{LIBERO-}\\\textbf{Long}\end{tabular} &
\textbf{Average} &
\textbf{$\Delta$ Avg.} \\ 
\midrule
\multirow{2}{*}{0.5B} & w/ motion & 84.0±0.9 & 86.6±0.9 & 78.0±1.1 & 46.0±1.3 & 73.7±0.6 & \multirow{2}{*}{+2.5} \\
                      & w/o motion & 84.8±0.9 & 85.8±0.9 & 69.6±1.2 & 44.4±1.3 & 71.2±0.6 & \\ 
\cmidrule{1-8} 
\multirow{2}{*}{1.5B} & w/ motion & 84.6±0.9 & 85.2±0.9 & 76.4±1.1 & 47.8±1.3 & 73.5±0.6 & \multirow{2}{*}{+1.8} \\
                      & w/o motion & 84.0±0.9 & 84.8±1.0 & 71.0±1.2 & 46.8±1.3 & 71.7±0.6 & \\ 
\cmidrule{1-8} 
\multirow{2}{*}{3B}   & w/ motion & 83.6±0.9 & \textbf{86.8±0.9} & \textbf{79.6±1.0} & \textbf{48.0±1.3} & \textbf{74.5±0.6} & \multirow{2}{*}{+2.5} \\
                      & w/o motion & \textbf{85.0±0.9} & 86.4±0.9 & 71.2±1.1 & 45.4±1.3 & 72.0±0.6 & \\ 
\bottomrule
\end{tabular}
\caption{Success rates (\%) on LIBERO benchmark (models trained from scratch)}
\label{table:scratchLibero}
\end{table}

\begin{table}[t]
\centering
\scriptsize
\renewcommand{\arraystretch}{1.0}
\setlength{\tabcolsep}{2.5pt}
\begin{tabular}{@{}cc|c|c|c|c|c|c@{}} 
\toprule
\multicolumn{2}{c|}{\textbf{From Scratch}} &
\begin{tabular}[c]{@{}c@{}}\textbf{Spoon/}\\\textbf{Towel}\end{tabular} &
\begin{tabular}[c]{@{}c@{}}\textbf{Carrot/}\\\textbf{Plate}\end{tabular} &
\begin{tabular}[c]{@{}c@{}}\textbf{Stack}\\ \textbf{Blocks} \end{tabular} &
\begin{tabular}[c]{@{}c@{}}\textbf{Eggplant/}\\\textbf{Basket}\end{tabular} &
\textbf{Average} &
\textbf{$\Delta$ Avg.} \\ 
\midrule
\multirow{2}{*}{0.5B} & w/ motion   & 16.7 & 19.6 & 0  & 16.4 & 13.2 & \multirow{2}{*}{+1.5} \\
                      & w/o motion  & 14.1 & 20.5 & 0  & 12.2  & 11.7  & \\ 
\cmidrule{1-8} 
\multirow{2}{*}{1.5B} & w/ motion   & 22.7 & 16.5 & 0  & 20.9 & 15.0 & \multirow{2}{*}{-0.3} \\
                      & w/o motion  & 33.1 & 19.8 & 0  & 8.6  & 15.3 & \\ 
\cmidrule{1-8} 
\multirow{2}{*}{3B}   & w/ motion   & \textbf{35.8} & \textbf{25.1} & 0  & \textbf{56.5} & \textbf{29.4} & \multirow{2}{*}{+9.6} \\
                      & w/o motion  & 28.5  & 24.4 & 0  & 26.3  & 19.8 & \\ 
\bottomrule
\end{tabular}
\caption{Success rates (\%) on SimplerEnv benchmark (models trained from scratch)}
\label{table:scratchSimpler}
\end{table}

To validate the effectiveness of our proposed pretraining strategy compared to other existing methods, we conducted comprehensive experiments across different model sizes. Specifically, we pretrained models solely focusing on motion generation and models focusing exclusively on action generation. Subsequent fine-tuning was performed across various benchmarks to evaluate their performance. Experimental results clearly demonstrate that models pretrained with our motion-focused method achieve significantly higher success rates (SR). Additionally, by comparing results from Table~\ref{table:scratchLibero}--\ref{table:pretrainSimpler}, it becomes evident that pretraining with motion provides a substantially greater improvement over baseline models trained from scratch. This further underscores the effectiveness and importance of incorporating motion learning during the pretraining stage. However, it was observed that the 1.5B parameter model exhibited limited performance gains in the SimplerEnv benchmark. We suggest this phenomenon arises primarily due to the gap between fine-tuning data, which was collected from real-world scenarios, and the simulated testing environment, compounded by the constraints imposed by the limited parameter scale.

Furthermore, we aimed to assess the efficacy of our two proposed methods for optimizing motion generation quality: adjusting the window size and threshold parameters. Experiments were conducted using the 0.5B parameter model on the LIBERO benchmark, comparing three scenarios—without motion pretraining, with original motion pretraining, and with our optimized motion pretraining. Results from Table~\ref{table:raw_motion} align well with human judgment assessments described in our methodology, demonstrating that our optimization techniques substantially enhance motion generation quality. This robust evidence confirms the significant contribution of our optimization strategies in improving pretraining outcomes.

\begin{table}[t]
\centering
\scriptsize 
\renewcommand{\arraystretch}{1.0}
\setlength{\tabcolsep}{1.5pt}
\begin{tabular}{cc|c|c|c|c|c|c} 
    \toprule
    \multicolumn{2}{c|}{\textbf{After Pretrain}} &
      \begin{tabular}[c]{@{}c@{}}\textbf{LIBERO-}\\\textbf{Spatial}\end{tabular} &
      \begin{tabular}[c]{@{}c@{}}\textbf{LIBERO-}\\\textbf{Object}\end{tabular} &
      \begin{tabular}[c]{@{}c@{}}\textbf{LIBERO-}\\\textbf{Goal}\end{tabular} &
      \begin{tabular}[c]{@{}c@{}}\textbf{LIBERO-}\\\textbf{Long}\end{tabular} &
      \textbf{Average} &
      \textbf{$\Delta$ Avg.} \\ 
    \midrule
    \multirow{2}{*}{0.5B}  & w/ motion  & 86.0±0.9 & 84.8±0.9 & 76.4±1.1 & 51.2±1.3 & 74.6±0.6 & \multirow{2}{*}{+3.2} \\
        & w/o motion   & 85.2±0.9 & 82.2±1.0 & 69.0±1.2 & 49.0±1.3 & 71.4±0.6 & \\ 
        \cmidrule{1-8} 
        \multirow{2}{*}{1.5B}  & w/ motion  & \textbf{86.8±0.9} & 84.8±0.9 & 75.2±1.1 & 51.6±1.3 & 74.6±0.6 & \multirow{2}{*}{+3.0} \\
        & w/o motion   & 84.4±0.9 & 82.8±1.0 & 69.8±1.2 & 49.4±1.3 & 71.6±0.6 & \\ 
        \cmidrule{1-8} 
        \multirow{2}{*}{3B}    & w/ motion  & 84.8±0.9 & \textbf{90.0±0.8} & \textbf{82.2±1.0} & \textbf{55.4±1.3} & \textbf{78.1±0.5} & \multirow{2}{*}{+6.9} \\
        & w/o motion   & 82.0±1.0 & 85.6±0.9 & 70.2±1.2 & 46.8±1.3 & 71.2±0.6 & \\ 
    \bottomrule
\end{tabular}
\caption{Success rates (\%) on LIBERO benchmark (models
trained based on pretraining)}
\label{table:pretrainLiberoNoBaseline}
\end{table}

\begin{table}[t]
\centering
\scriptsize 
\renewcommand{\arraystretch}{1.0}
\setlength{\tabcolsep}{1.5pt}
\begin{tabular}{cc|c|c|c|c|c|c}
    \toprule
    \multicolumn{2}{c|}{\textbf{After Pretrain}} &
      \begin{tabular}[c]{@{}c@{}}\textbf{Spoon /}\\ \textbf{Towel}\end{tabular} &
      \begin{tabular}[c]{@{}c@{}}\textbf{Carrot /}\\ \textbf{Plate}\end{tabular} &
      \begin{tabular}[c]{@{}c@{}}\textbf{Stack} \\\textbf{Blocks} \end{tabular} &
      \begin{tabular}[c]{@{}c@{}}\textbf{Eggplant/ }\\ \textbf{Basket}\end{tabular} &
      \textbf{Average} &
      \textbf{$\Delta$ Avg.} \\ 
    \midrule
    \multirow{2}{*}{0.5B}  & w/ motion  & 21.1  & 24.4  & 0.0   & 10.7  & 14.1 & \multirow{2}{*}{+2.3} \\
        & w/o motion   & 18.5  & 12.4  & 0.0   & 16.3   & 11.8 & \\ 
        \cmidrule{1-8} 
        \multirow{2}{*}{1.5B}  & w/ motion  & 30.8  & 20.8  & 0.0   & 24.4  & 19.0 & \multirow{2}{*}{+0.8} \\
        & w/o motion   & 34.6  & 22.1  & 0.0   & 16.2  & 18.2 & \\ 
        \cmidrule{1-8} 
        \multirow{2}{*}{3B}    & w/ motion  & \textbf{44.0}  & \textbf{36.2}  & 0.0   & \textbf{61.1}  & \textbf{35.3} & \multirow{2}{*}{+14.1} \\
        & w/o motion   & 26.4  & 28.4  & 0.0   & 30.1  & 21.2 & \\ 
    \bottomrule
\end{tabular}
\caption{Success rates (\%) on SimplerEnv benchmark (models
trained based on pretraining)}
\label{table:pretrainSimpler}
\end{table}

\begin{table}
\scriptsize
\centering
\renewcommand{\arraystretch}{1.0}
\setlength{\tabcolsep}{1.5pt}
\begin{tabular}{c|c|c|c|c|c}
    \toprule
    \textbf{Method} &
    \begin{tabular}[c]{@{}c@{}}\textbf{LIBERO-}\\\textbf{Spatial}\end{tabular} &
    \begin{tabular}[c]{@{}c@{}}\textbf{LIBERO-}\\\textbf{Object}\end{tabular} &
    \begin{tabular}[c]{@{}c@{}}\textbf{LIBERO-}\\\textbf{Goal}\end{tabular} &
    \begin{tabular}[c]{@{}c@{}}\textbf{LIBERO-}\\\textbf{Long}\end{tabular} &
    \textbf{Average} \\
    \midrule
    Ours (0.5B) &
    86.2±0.9 &
    84.6±0.9 &
    76.2±1.1 &
    51.1±1.3 &
    74.5±0.6 \\
    
    with raw motion &
    86.0±0.9 &
    83.2±1.0 &
    74.1±1.1 &
    50.5±1.3 &
    73.5±0.6 \\
    
    w/o motion &
    85.1±0.9 &
    82.3±1.0 &
    69.0±1.2 &
    49.3±1.3 &
    71.4±0.6 \\
    \bottomrule
\end{tabular}
\caption{Success rates (\%) on Ours and raw motion and without motion}
\label{table:raw_motion}
\end{table}

\subsection{Does our refined model outperform baselines and SOTA? (RQ2)}
\begin{table}[t]
\centering
\scriptsize 
\renewcommand{\arraystretch}{1.0}
\setlength{\tabcolsep}{1.5pt}

\begin{tabular}{cc|c|c|c|c|c}
    \toprule
    \multicolumn{2}{c|}{\textbf{Method}} &
      \begin{tabular}[c]{@{}c@{}}\textbf{LIBERO-}\\\textbf{Spatial}\end{tabular} &
      \begin{tabular}[c]{@{}c@{}}\textbf{LIBERO-}\\\textbf{Object}\end{tabular} &
      \begin{tabular}[c]{@{}c@{}}\textbf{LIBERO-}\\\textbf{Goal}\end{tabular} &
      \begin{tabular}[c]{@{}c@{}}\textbf{LIBERO-}\\\textbf{Long}\end{tabular} &
      \textbf{Average} \\
    \midrule
    \multicolumn{2}{c|}{Diffusion Policy} & 78.3±1.1 & \textbf{92.5±0.7} & 68.3±1.2 & 50.5±1.3 & 72.4±0.7 \\
    \multicolumn{2}{c|}{ScaleDP}          & 79.1±0.7 & 90.4±0.9          & 73.6±0.8 & 48.4±1.2 & 72.9±0.5 \\
    \multicolumn{2}{c|}{Octo}             & 78.9±1.0 & 85.7±0.9          & \textbf{84.6±0.9} & 51.1±1.3 & 75.1±0.6 \\
    \multicolumn{2}{c|}{Openvla}      & 84.7±0.9 & 88.4±0.8          & 79.2±1.0 & 53.7±1.3 & 76.5±0.6 \\
    \midrule
    \multirow{3}{*}{Ours} 
        & 0.5B & 86.0±0.9 & 84.8±0.9 & 76.4±1.1 & 51.2±1.3 & 74.6±0.6 \\
        \cmidrule{2-7}
        & 1.5B & \textbf{86.8±0.9} & 84.8±0.9 & 75.2±1.1 & 51.6±1.3 & 74.6±0.6 \\
        \cmidrule{2-7}
        & 3B   & 84.8±0.9 & 90.0±0.8 & 82.2±1.0 & \textbf{55.4±1.3} & \textbf{78.1±0.5} \\
    \bottomrule
\end{tabular}

\caption{Success rates (\%) on LIBERO benchmark (models
trained based on pretraining)}
\label{table:pretrainLibero}
\end{table}


We compared our method with state-of-the-art manipulation baselines. Unlike these baselines, our objective explicitly aligns motion components—the aspect most sensitive to dataset-driven numeric shifts. This simplification accelerates convergence and yields superior results on two benchmarks, proving the approach captures transferable motion directions with modest compute.
\begin{table}[t]
\centering
\scriptsize 
\renewcommand{\arraystretch}{1.0}
\setlength{\tabcolsep}{6pt}

\begin{tabular}{cc|c|c|c|c|c}
    \toprule
    \multicolumn{2}{c|}{\textbf{Method}} &
      \begin{tabular}[c]{@{}c@{}}\textbf{Spoon /}\\ \textbf{Towel}\end{tabular} &
      \begin{tabular}[c]{@{}c@{}}\textbf{Carrot /}\\ \textbf{Plate}\end{tabular} &
      \begin{tabular}[c]{@{}c@{}}\textbf{Stack} \\\textbf{Blocks} \end{tabular} &
      \begin{tabular}[c]{@{}c@{}}\textbf{Eggplant/} \\ \textbf{Basket}\end{tabular} &
      \textbf{Average} \\
    \midrule
    \multicolumn{2}{c|}{RT1-x}        & 0.0  & 4.2  & 0.0  & 0.0  & 1.1  \\
    \multicolumn{2}{c|}{Octo-Base}    & 15.8 & 12.5 & 0.0  & 41.7 & 17.5 \\
    \multicolumn{2}{c|}{Octo-Small}   & 41.7 & 8.2  & 0.0  & 56.7 & 26.7 \\
    \multicolumn{2}{c|}{Openvla}      & 4.2  & 0.0  & 0.0  & 12.5 & 4.2  \\
    \multicolumn{2}{c|}{ECoT}         & 40.2 & 11.7 & 0.0  & 28.4 & 20.1 \\
    \midrule
    \multirow{3}{*}{Ours} 
        & 0.5B & 21.1  & 24.4  & 0.0   & 10.7  & 14.1 \\
        \cmidrule{2-7}
        & 1.5B & 30.8  & 20.8  & 0.0   & 24.4  & 19.0 \\
        \cmidrule{2-7}
        & 3B   & \textbf{44.0}  & \textbf{36.2}  & 0.0   & \textbf{61.1}  & \textbf{35.3} \\
    \bottomrule
\end{tabular}

\caption{Success rates (\%) on SimplerEnv benchmark (models
trained based on pretraining)}
\label{table:pretrainSimplerNoBaseline}
\end{table}

Specifically, we fine-tuned the pretrained models and evaluated them on the LIBERO and SimplerEnv benchmarks. As shown in Table~\ref{table:pretrainLibero} and Table~\ref{table:pretrainSimplerNoBaseline}, our approach consistently surpasses baselines lacking motion tokens, and it also outperforms models trained entirely from scratch. We further compared ECoT (7B), trained on the same Bridge dataset, and found our method still attained superior performance. Notably, the OpenVLA (7B) variant of our approach delivered particularly competitive results while requiring fewer pretraining data and smaller model sizes.

All models failed to succeed in the task ``stack green block on yellow block" in SimplerEnv, This may be because it requires a higher level of precision than other tasks, because it first needs to grab a very small block and put it accurately on another very small block. Of course, we think the main reason is that the training data used is Bridge v2, which is collected from the real world, and our test environment is SimplerEnv, which is a test environment replicated in simulation based on the Bridge v2 dataset, which is different from the actual environment.

\subsection{Does a language output objective reduce action-language distance? (RQ3)}
\begin{figure}[t] 
    \centering
    \begin{subfigure}[b]{0.45\linewidth} 
        \centering
        \includegraphics[width=\linewidth]{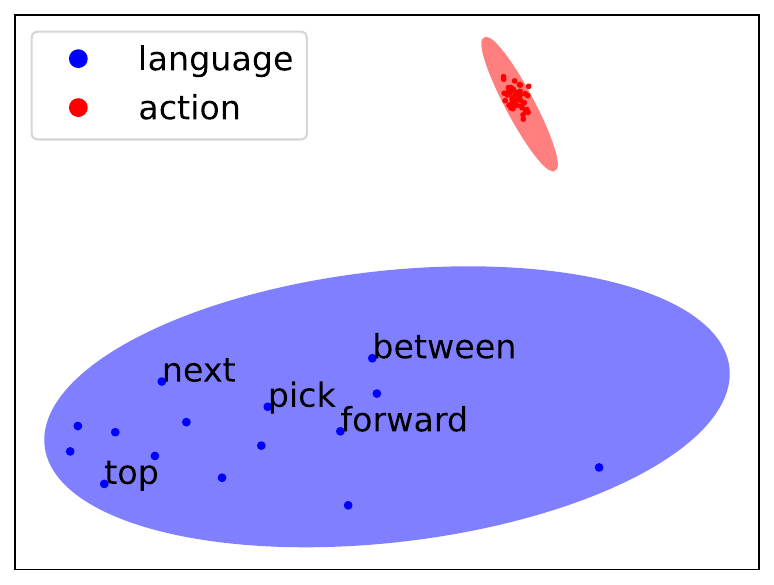}
        \caption{Baseline from Scratch}
        \label{fig:baseline_scratch}
    \end{subfigure}
    \hfill 
    \begin{subfigure}[b]{0.45\linewidth}
        \centering
        \includegraphics[width=\linewidth]{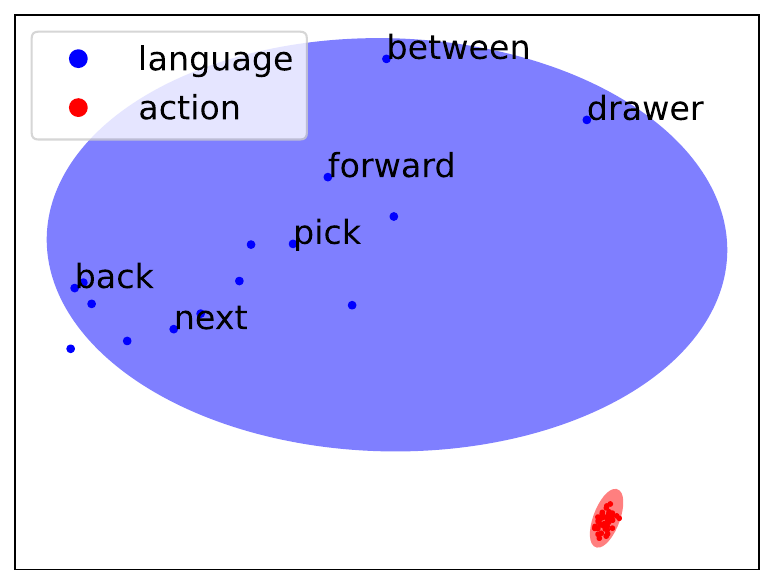}
        \caption{Baseline after Pretraining}
        \label{fig:baseline_pretrain}
    \end{subfigure}
    
    \vspace{0.5em}

    \begin{subfigure}[b]{0.45\linewidth}
        \centering
        \includegraphics[width=\linewidth]{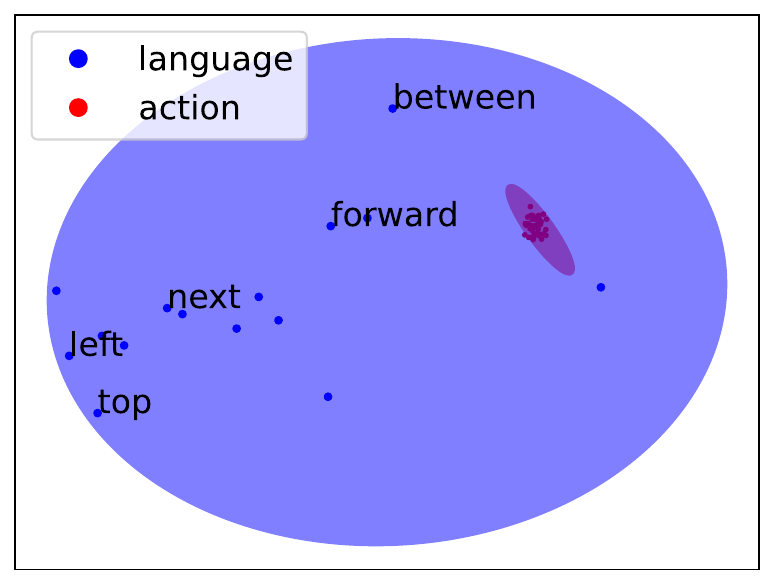}
        \caption{Ours from Scratch}
        \label{fig:ours_scratch}
    \end{subfigure}
    \hfill
    \begin{subfigure}[b]{0.45\linewidth}
        \centering
        \includegraphics[width=\linewidth]{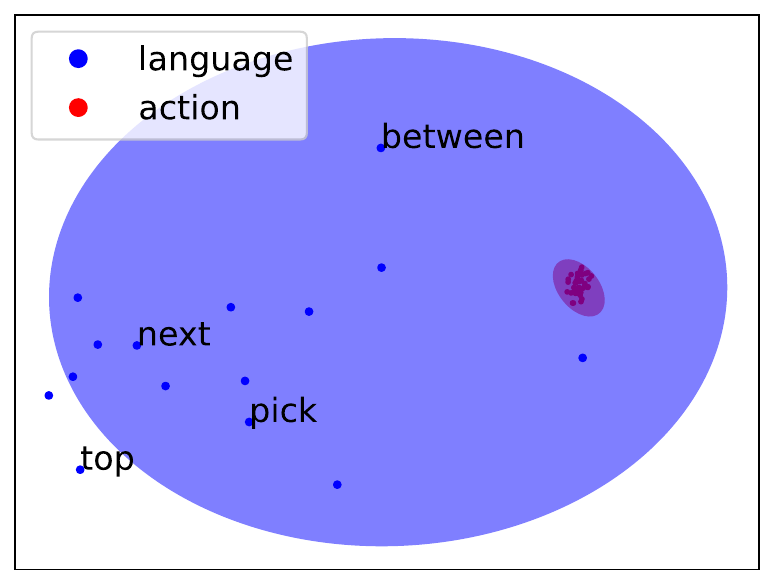}
        \caption{Ours after Pretraining}
        \label{fig:ours_pretrain}
    \end{subfigure}
    
    \caption{Comparison of four different experimental setups}
    \label{fig:comparison}
    \vspace{-10pt}
\end{figure}

Building on the findings from RQ2, we further investigate the effect of incorporating motion tokens on representation alignment. Specifically, we leverage PCA and confidence ellipses to visualize embeddings from the spatial task of the LIBERO benchmark(see Fig.~\ref{fig:comparison}). We selected one of the train-sets on the spatial task, extracted the embeddings of the model under four training conditions (w\&w/o pretraining; w\&w/o motion), and then took the vectors corresponding to action tokens and motion tokens in the vocabulary from the embeddings, used principal component analysis to reduce the dimensionality, and then drew its confidence ellipse and marked the relevant tokens at the corresponding points.

Visualizations indicate that in end-to-end models trained for action token generation, the token features deviate significantly from those of the original vocabulary. Incorporating our motion representation (whether pretrained or trained from scratch) reduces this gap, helping to bridge the modality difference \citep{wei2025prism} and leading to more efficient training. Moreover, pretraining yields more clustered action token features, coinciding with improved manipulation performance, whereas training from scratch results in more dispersed representations, suggesting insufficient convergence.

\section{Conclusion}
In this paper, we introduce a novel pretraining strategy using language-modal action representations (motion) to tackle generalization issues caused by numerical distribution shifts across robotic platforms and tasks. Our method converts numerical actions into abstract directional semantic descriptions, significantly reducing distributional discrepancies and enabling efficient autonomous learning of generalized motion-language alignments. Experiments confirm that integrating motion tokens effectively bridges representational gaps between action and language modalities, enhancing generalization and transferability across various robotic manipulation benchmarks, all without external modules or manual intervention. Future work will focus on further optimizing the pretraining strategy to advance language-guided robotic manipulation toward practical applications.

\section{Acknowledgments}
The authors would like to thank all the anonymous reviewers for their insightful comments. Meanwhile, the authors would like to thank Prof. Wei-Nan Zhang and Dr. Yuanxing Liu for their help on revising the manuscript. 
This research was supported by the National Key Research and Development Program (No.2022YFF0902100), the National Natural Science Foundation of China (No. 92470205) and the National Natural Science Foundation of China (No. 20230320).

\bibliography{aaai2026}

\end{document}